\documentclass[letterpaper, 10 pt, conference]{ieeeconf}  % Comment this line out if you need a4paper

\IEEEoverridecommandlockouts                              % This command is only needed if 
\usepackage{amssymb}  % assumes amsmath package installed
\usepackage{graphicx}
\newcommand{\hide}[1]{}

\title{\LARGE \bf
Self-Supervised Attention Learning for Depth and Ego-motion Estimation}
\author{Assem Sadek and Boris Chidlovskii \\
Naver Labs Europe, chemin Maupertuis 6, Meylan-38240, France \\
        {\tt\small firstname.lastname@naverlabs.com}% <-this % stops a space
}

\begin{document}

\maketitle
\thispagestyle{empty}
\pagestyle{empty}

%%%%%%%%%%%%%%%%%%%%%%%%%%%%%%%%%%%%%%%%%%%%%%%%%%%%%%%%%%%%%%%%%%%%%%%%%%%%%%%%
\begin{abstract}
We address the problem of depth and ego-motion estimation from image sequences. Recent advances in the domain propose to train a deep learning model for both tasks using image reconstruction in a self-supervised manner. We revise the assumptions and the limitations of the current approaches and propose two improvements to boost the performance of the depth and ego-motion estimation. We first use Lie group properties to enforce the geometric consistency between images in the sequence and their reconstructions. We then propose a mechanism to pay attention to image regions where the image reconstruction gets corrupted. We show how to integrate the attention mechanism in the form of attention gates in the pipeline and use attention coefficients as a mask.
We evaluate the new architecture on the KITTI datasets and compare it to the previous techniques. We show that our approach improves the state-of-the-art results for ego-motion estimation and achieve comparable results for depth estimation.
\end{abstract}

%%%%%%%%%%%%%%%%%%%%%%%%%%%%%%%%%%%%%%%%%%%%%%%%%%%%%%%%%%%%%%%%%%%%%%%%%%%%%%%%

\section{INTRODUCTION}
\label{sec:intro}

The tasks of depth estimation and ego-motion are longstanding problems in computer vision; their successful solution is crucial for a wide variety of applications, such as autonomous driving, robot navigation, and visual localization, Augmented/Virtual Reality applications, etc.

In the last years, deep learning networks~\cite{eigen2015predicting,Garg2016UnsupervisedRescue,Liu2016LearningFields,zhou2017unsupervised} achieved results comparable with traditional geometric methods for depth estimation. They show competitive results in complex and ambiguous 3D areas, with CNNs serving as {\it deep regressors} and coupled with classical components, to get the best from the geometric and learning paradigms. For the ego-motion estimation, several works~\cite{Kendall2015PoseNet:Relocalization,zhou2017unsupervised}  have achieved a level of performance comparable to the traditional techniques based on the SLAM algorithm~\cite{Gomez2018VisualCase,Mur-Artal2015ORB-SLAM:System,Mur-Artal2017ORB-SLAM2:Cameras}. Early methods for depth and ego-motion (DEM) are based on supervised learning;  they require large annotated datasets and calibrated setups. Trained and tested on publicly available benchmark datasets, these techniques show a limited capacity to generalize beyond the data they are trained on. 

Moreover, data annotation is often slow and costly. The annotations also suffer from the structural artifacts, particularly in the presence of reflective, transparent, dark surfaces or non-reflective sensors that output infinity values. All these challenges strongly motivated the shift to the unsupervised learning of depth and ego-motion, in particular from monocular (single-camera) 
videos. 

To enable the DEM estimation without annotations, the major idea is to process both tasks jointly~\cite{zhou2017unsupervised}. In the self-supervised setting, an assumption is made about spatial consistency and temporal coherence between consecutive frames in a sequence. The only external data needed is the camera intrinsics. Recent progress in the domain~\cite{mahjourian2018unsupervised,xu2018structured,wang19icra,Bian19nips} allows us to use monocular unlabeled videos to provide self-supervision signals to a learning component. The 3D geometry estimation includes per-pixel depth estimation from a single image and 6DoF relative pose estimation between neighbor images.

Self-supervised learning greatly boosted DEM estimation performance. There however remains a gap with the related supervised methods. The underlying assumption of the static world is often violated in real scenes and the geometric image reconstruction gets corrupted by unidentified 
moving objects, occlusions, reflection effects, etc.

Multiple improvements have been recently proposed to address these issues~\cite{almalioglu2018ganvo,Bian19nips,mahjourian2018unsupervised,wang19icra,wang2018joint}. They are often based on adding more components to the architecture such as flow nets~\cite{wang2018joint}, semantic segmentation~\cite{mahjourian2018unsupervised}, adversarial networks~\cite{almalioglu2018ganvo}, and multiple masks~\cite{wang19icra}. These approaches lead however to an important growth of model parameters, making the architecture and training procedure more complex.

In this paper, we propose an alternative and effective solution to the
problem-based on the {\it attention} mechanism~\cite{jetley18learn}. Initially proposed for natural language processing tasks~\cite{chaudhari2019attentive}, the attention and its variants have been successfully extended to computer vision tasks, including image classification~\cite{wang17residual}, semantic segmentation~\cite{huang2017,oktay2019}, image captioning
~\cite{xu15show}, and depth estimation ~\cite{xu2018structured}. Inspired by these successes, we propose to include the attention mechanism in self-supervised learning for DEM estimation. We show that so-called attention gates can be integrated into the baseline architecture and trained from scratch to automatically learn to focus on corrupted regions without additional supervision.

The attention gates do not require a large number of model parameters and introduce a minimal computational overhead. In return, the proposed mechanism improves model sensitivity and accuracy for dense depth and ego-motion estimation.

The attention gates are integrated into the depth estimation network. Consequently, the depth network can predict both the depth estimation
and attention coefficients which are then used to weigh the difference between the true and reconstructed pixels when minimizing the objective function. 

We evaluate the proposed architecture on the KITTI datasets and compare it to the state of the art techniques. We show that our approach improves the state-of-the-art results for ego-motion estimation and achieve comparable results for depth estimation.

%------- the architecture diagram ------
\begin{figure*}[ht]
\centering
  \includegraphics[width=\linewidth]{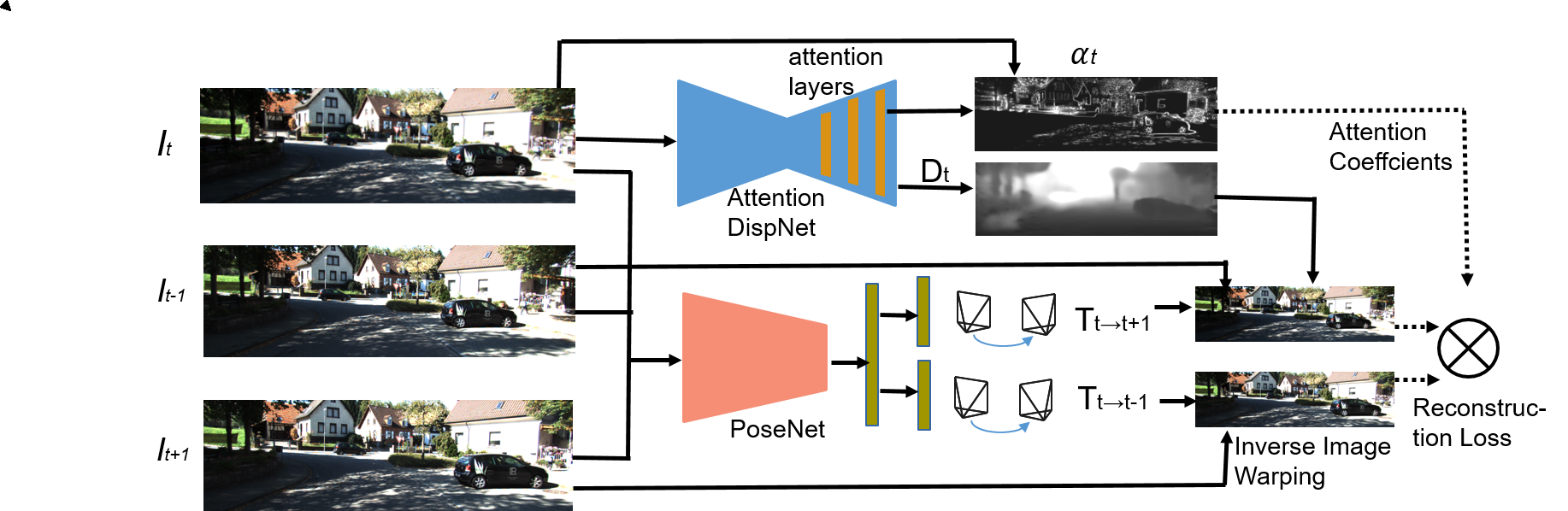}
  \caption{Attention-based DEM architecture.}
  \label{fig:arch}
\end{figure*}

%---------------------------------------
\section{RELATED WORK}
\label{sec:soa}

Eigen at al.~\cite{eigen2014depth} was first to directly regress a CNN over pixel values and to use multi-scale features for monocular depth estimation. They used the global (coarse-scale) and local (fine-scale) networks to accomplish the tasks of global depth prediction and local refinements. 

Garg et al.~\cite{Garg2016UnsupervisedRescue} proposed to use a calibrated stereo camera pair setup where the depth is produced as an intermediate output and the supervision comes as a reconstruction of one image from another in a stereo pair. Images on the stereo rig have a fixed and known transformation, and the depth can be learned from this functional relationship.

An important step forward was developed in Godard et al.~\cite{godard2016unsupervised} where the depth estimation problem was reformulated in a new way. Godard et al. employ binocular stereo pairs of a view in training but, during inference time, one view is only used to estimate the depth. By exploiting epipolar geometry constraints, they generate disparity images by training their network with an image reconstruction loss. The model does not require any labeled depth data and learns to predict pixel-level correspondences between pairs of rectified stereo images. 

Mahjourian et al.~\cite{mahjourian2018unsupervised} made another step by using camera ego-motion and 3D geometric constraints. Zhou et al.~\cite{zhou2017unsupervised} proposed a novel approach for unsupervised learning of depth and ego-motion from monocular video only. An additional module to learn the motion of objects was introduced in~\cite{Vijayanarasimhan2017}; however, their architecture recommends optional supervision by ground-truth depth or optical flow to improve performance.

%================== Last improvements ===============================
The static world assumption doe not hold in real scenes, because of unidentified moving objects, occlusions, photo-effects, etc. that violate the underlying assumption and corrupt the geometric image reconstruction. The recent works address these limitations and propose several improvements, varying from new objective functions, additional modules and pixel masking to new learning schemes.

Almalioglu at al.~\cite{almalioglu2018ganvo} proposed a framework that predicts pose camera motion and a monocular depth map of the scene using deep convolutional Generative Adversarial Networks (GANs). An additional adversarial module helps learn more accurate models and make reconstructed images indistinguishable from the real images.

%------------ICRA 19---------------
Wang et al.~\cite{wang19icra} coped with errors in realistic scenes due to reflective surfaces and occlusions. They combined the geometric and photometric losses by introducing the matching loss constrained by epipolar geometry and designed multiple masks to solve image pixel mismatch caused by the movement of the camera. 

%-------------------------------------------
Another solution was proposed in UnDEMoN architecture~\cite{babu18iros}. The authors changed the objective function and tried to minimize spatial and temporal reconstruction losses simultaneously. These losses are defined using a bi-linear sampling kernel and penalized using the Charbonnier penalty function. 

%----------Nips19 by Bian-------------
Most recently, Bian et al.\cite{Bian19nips} analyzed violations of the underlying static assumption in geometric image reconstruction and concluded that, due to lack of proper constraints, networks output scale-inconsistent results over different samples. To remedy the problem, the authors proposed a geometry consistency loss for scale-consistent predictions and a mask for handling moving objects and occlusions. Since our approach does not leverage additional modules nor multi-task learning, our attention-based framework is much simpler and more efficient. 
\section{Baseline architecture and extensions}
\label{sec:architecture}
Similar to the recent methods~\cite{mahjourian2018unsupervised,wang19icra,zhou2017unsupervised}, our baseline architecture includes depth estimation and pose estimation modules. The depth module is an encoder-decoder network (DispNet); it takes a target image and outputs depth values $\hat{D}_t(p)$ for every pixel $p$ in the image. The pose module (PoseNet) take as input the concatenation of the target image $I_t$ and two neighbors (source) images $I_s$, $s\in \{t-1,t+1\}$. It outputs transformation matrices $\hat{T}_{t\rightarrow s}$, representing the six degrees of freedom (6DoF) relative pose between the images.

%----INVERSE WARPING OPERATION-----------------
\subsubsection{Image reconstruction}
Self-supervised learning is proceeded by image reconstruction using the {\it inverse warping} technique~\cite{godard2016unsupervised}. This technique is differentiable, therefore it allows us to back-propagate the gradients during training. It tries to reconstruct the target image  $I_{t}$ through sampling pixels from the source images $I_s$ based on the estimated depth map $\hat{D}_t$ and the relative pose transformation matrices $\hat{T}_{t\rightarrow s}$, $s \in \{t-1,t+1\}$. The sampling is done by projecting the homogeneous coordinates of the target pixel $p_t$ onto the source view $p_s$. Given the camera intrinsics $K$, the estimated depth of the pixel $\hat{D}_t(p)$ and transformation matrix $\hat{T}_{t\rightarrow s}$, the projection is done by the following equation.
%-------------------------
\begin{equation}
    p_s \sim K \hat{T}_{t\rightarrow s}\hat{D}_{t}(p_t)K^{-1}p_t.
    \label{eq:inverse_warping}
\end{equation}
%------------------------
For non-discrete values of a projected pixel position, the {\it differentiable bi-linear sampling interpolation}~\cite{Jaderberg2015SpatialNetworks} is used to find the intensity value at a given position. The mean intensity value in the reconstructed image $\hat{I_s}$ is interpolated using 4-pixel neighbors of $p_s$ ({\it t}op-{\it r}ight, {\it t}op-{\it l}eft, {\it b}ottom-{\it r}ight, {\it b}ottom-{\it l}eft), as follows
%-------------------------
\begin{equation}
    \hat{I}_s(p_t) = I_s(p_s) = \sum_{i\in\{t,b\}, j\in\{l,r\}}w^{ij}I_s(p_s^{ij}),
    \label{eq:interpolate}
\end{equation}
%-------------------------
\noindent
Where $\hat{I}_s(p_t)$ is the intensity value of $p_t$ in the reconstructed image $\hat{I_s}$. The weight is linearly proportional to the spatial proximity between $p_s$ and the neighbor $p_s^{ij}$; the four weights $w^{ij}$ sum to 1.
%=======================================
\subsubsection{Photometric Loss}
Under the static world assumption, many existing methods apply the {\it photometric} loss~\cite{almalioglu2018ganvo,shen19icra} defined as $L_1$ loss objective function: $\mathcal{L}_{p} = \sum_s \sum_p | I_t (p)-\hat{I}_s(p)|$.

Any violation of the static world assumption in the real scenes affects drastically the reconstruction. 
To overcome this limitation, one solution is to use the SSID loss~\cite{mahjourian2018unsupervised,wang19icra,shen19icra}. A more advanced solution~\cite{zhou2017unsupervised} is to introduce an \textit{explainability} mask to indicate the importance of a pixel in the warped images. If the pixel contributes to a corrupted synthesis 
the explainability value of the pixel will be negligible. 

The explainability values are produced by a dedicated module (ExpNet) in~\cite{zhou2017unsupervised}; it shares the encoder with the PoseNet and branches off in the decoding part. The three networks(DispNet, PoseNet and ExpNet) are trained simultaneously. The ExpNet decoder generates a per-pixel mask $\hat{E}_k(p)$. Similar to PoseNet, the explainability map $\hat{E}_k$  is generated for both source images. Per-pixel explainability values are embedded in the photometric loss:
%----------------
\begin{equation}
    \mathcal{L}_{p} = \frac{1}{|V|}\sum_p \hat{E}_k(p) | I_t(p) - \hat{I}_s(p) |~.
\label{eq:exp}
\end{equation}
%-----------------
where $|V|$ is the number of pixels in the image.
To avoid a trivial solution in (\ref{eq:exp}) with $\hat{E}_k(p)$ equals to zero, a constraint is added on the values of $\hat{E}_k(p)$. This constraint is implemented as a regularization loss $\mathcal{L}_{reg}(\hat{E}_k)$, defined as a cross entropy loss between the mask value and a constant 1.
%--------------------------------------------
\subsubsection{Depth smoothness}
We follow~\cite{shen19icra} in including a smoothness term to 
resolve the gradient-locality issue and remove discontinuity of the learned depth in low-texture regions. We use the edge-aware depth smoothness loss which uses image gradient to weigh the depth gradient:
\begin{equation}
L_{smo} = \sum_p|\nabla D(p)|^{T} \cdot e^{-|\nabla I(p)|},
\label{eq:smooth}
\end{equation}
\noindent
where $p$ is the pixel on the depth map $D$ and image $I$, $\nabla$ denotes the 2D differential operator, and $|\cdot|$ is the element-wise absolute value. We apply the smoothness loss on three additional intermediate layers from DispNet.

\begin{figure}[ht]
\centering
    \includegraphics[width=0.7\linewidth]{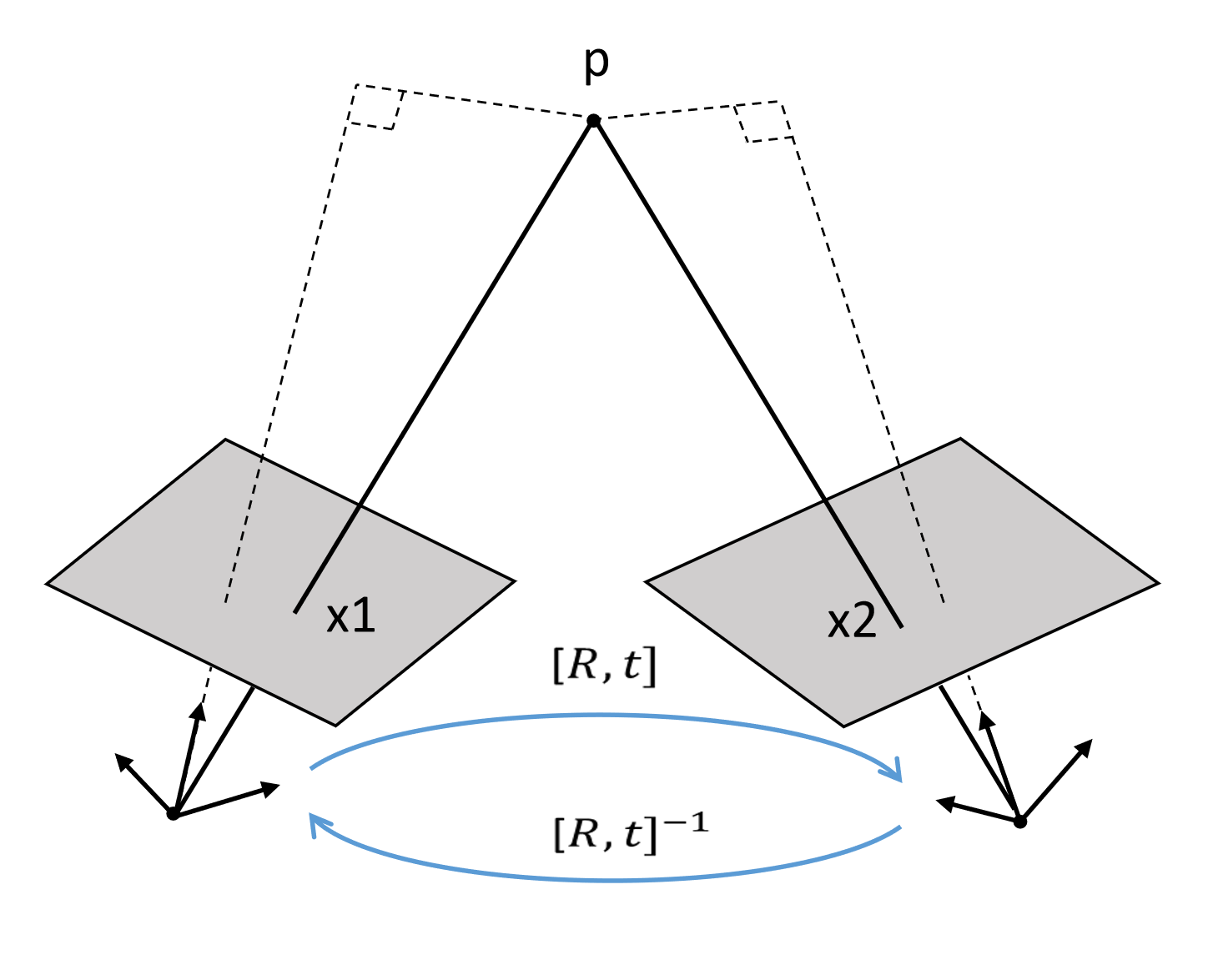}
    \caption{Relative transformation between two different views for the same camera.}
  \label{fig:BF}
\end{figure}

%=========================================================
\subsection{Backward-Forward Consistency} % Loss (Geometry-based approach)}
\label{ssec:BF}

The baseline architecture presented in the previous section integrates all components that proved their efficiency in the state of the art methods. Now we propose the first extension of our architecture and consider reinforcing the geometric consistency by using the Lie group property~\cite{claraco19}.

Indeed, the set of 3D space transformations $T$ form a Lie group $\mathbb{SE}(3)$; it is represented by linear transformations on homogeneous vectors, $T =[{\bf R,t}] \in \mathbb{SE}(3)$, with the rotation component $\mathbf{R} \in \mathbb{SO}(3),$ and translation component $\:\mathbf{t}\in\mathbb{R}^{3}$. For every transformation $T \in \mathbb{SE}(3)$, there is an inverse transformation $T^{-1} \in \mathbb{SE}(3)$, such that $T T^{-1} = I$ (see Figure~\ref{fig:BF}).

The PoseNet estimates relative pose transformations from a given target to the source frames. Therefore, for every pair of neighbour frames $(t-1,t)$ or $(t,t+1)$, we obtain the forward transformation $\hat{T}_{t-1\rightarrow t}$ as well as the backward one $\hat{T}_{t\rightarrow t-1}$. We can get the two transformations by changing the order of the concatenation images.

In a general case, for every pair of transformations $\hat{T}_{t\rightarrow s}$ and $\hat{T}_{s\rightarrow t}$, we impose an additional {\it forward-backward} geometric constraint; it requires the product of forward and backward transformations to be as close as possible to the identity matrix $I_{4x4} \in \mathbb{SE}(3)$. The corresponding loss is defined over for all pairs of relative pose transformations:

\begin{equation}
    \mathcal{L}_{bf} = \sum_s \sum_t |\hat{T}_{s \rightarrow t}\hat{T}_{t\rightarrow s} - I_{4x4} |~.
\label{eq:BF}
\end{equation}

%===========================
The total training loss is given by
\begin{equation}
    \mathcal{L}_{total} = \mathcal{L}_{p} +  \lambda_{smo} \mathcal{L}_{smo} + \lambda_{reg} \mathcal{L}_{reg} + \lambda_{bf} \mathcal{L}_{bf},
\label{eq:full}
\end{equation}
where $\lambda_{smo}$, $\lambda_{reg}$, $\lambda_{bf}$ are hyper-parameters.
In the experiments, $\lambda_{smo}=0.1$, $\lambda_{reg}=0.1$ and $\lambda_{bf}=0.1$, as showing the most stable results.

%===========================================================
\subsection{Self-attention gates}
\label{ssec:attention}

Our second extension of the baseline architecture addresses the attention mechanism and lets the network know where to look as it is performing the task of DEM estimation. 

Unlike integrating attention in Conditional Random Fields~\cite{xu2018structured}, our proposal is inspired by the recent works in semantic segmentation~\cite{jetley18learn} in particular in medical imaging~\cite{oktay2019, schlemper2018}. We treat attention in-depth estimation similarly to semantic segmentation. If we consider that each instance (a group of pixels) belongs to a certain semantic label (e.g. pedestrian), then the same group of pixels will have close and discontinuous depth values. Hence, we pay attention to any violation of this principle as a potential source of corrupted image reconstruction.

\begin{figure}[h]
\centering
  \includegraphics[width=0.8\linewidth]{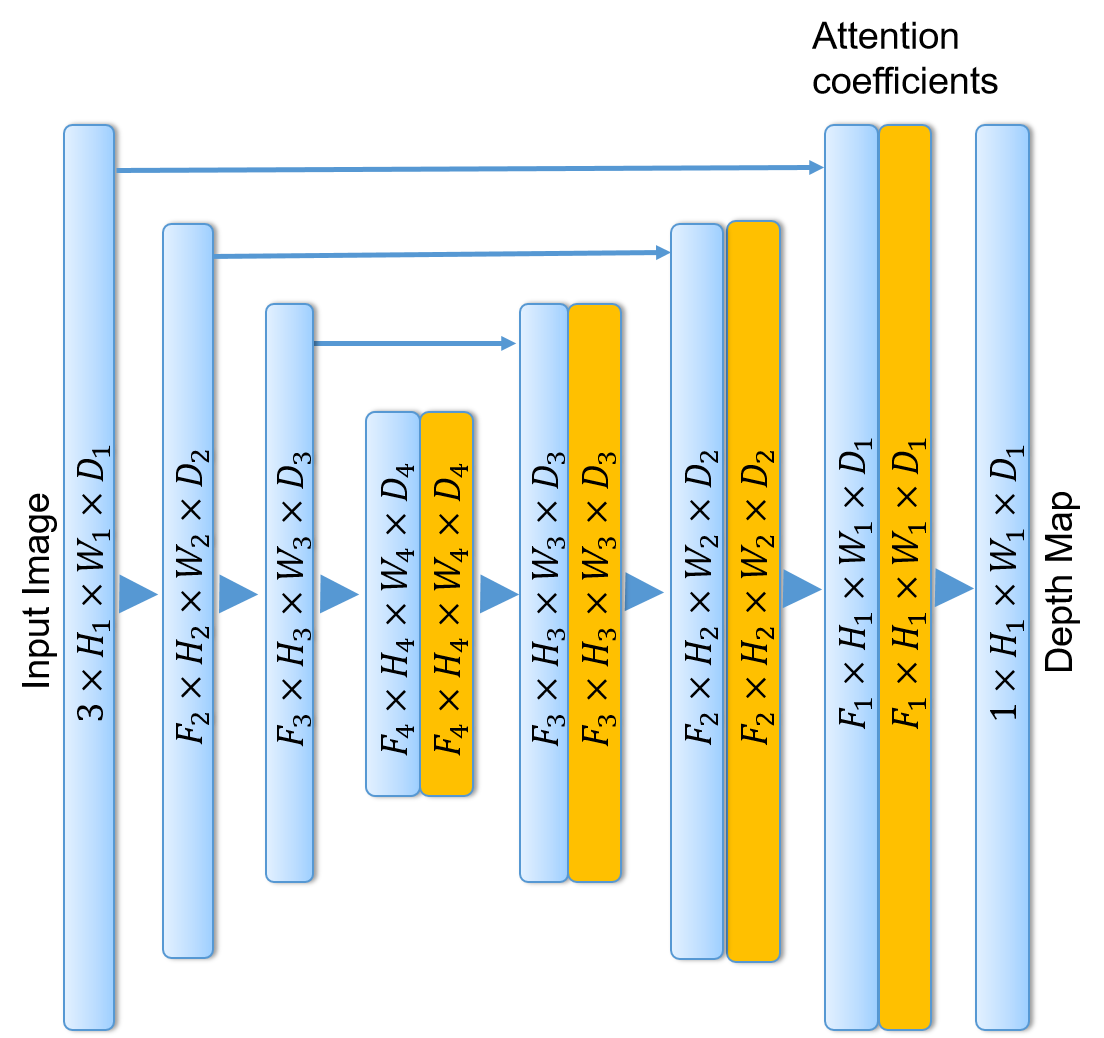}
  \caption{DispNet with the integrated attention gates (in orange). The input image is progressively filtered and downsampled by a factor of 2 at each scale $l$ in the encoding part of the network, $H_i=H_1/2^{i-1}$. Attention gates filter the features propagated through the skip connections by using the contextual information (gating) extracted in coarser scales.}
  \label{fig:attention}
\end{figure}

We propose to integrate the attention mechanism in the depth module (DispNet). As shown in Figure~\ref{fig:attention}, the encoder does not change, while the decoder layers are interleaved with the attention gates (AGs). The integration is done as follows.

Let ${\bf x}^l=\{{\bf x}_i^l\}^n_{i=1}$ be the activation map of a chosen layer $l \in \{1,\ldots,L\}$, where each ${\bf x}_i^l$ represents the pixel-wise feature vector of length $F_l$ (i.e. the number of channels). For each ${\bf x}_i^l$, AG computes coefficients $\alpha^l=\{\alpha_i^l\}^n_{i=1}$, $\alpha_i^l \in [0,1]$ to identify corrupted image regions, also, to prune feature responses that preserve only the activations relevant to the accurate depth estimation. The output of AG is ${\bf \hat{x}}^l= \{\alpha_i^l {\bf x}_i^l\}^n_{i=1}$ where each feature vector is scaled by the corresponding attention coefficient.

The attention coefficients $\alpha_i^l$ are computed as follows. In DispNet, the features on the coarse level identify the location of the target objects and model their relationship on a global scale. Let ${\bf g} \in \mathbb{R}^{F_{ge}}$ be such a global (coarser) feature vector providing information to AGs to disambiguate task-irrelevant feature content in ${\bf x}^l$. The idea is to consider each ${\bf x}^l$ and $\bf g$ jointly to attend the features at each scale $l$ that are most relevant to the objective being minimized.

The gating vector  ${\bf g}$ contains contextual information to prune lower-level feature responses ${\bf x}_i^l$ as suggested in AGs for image classification~\cite{wang17residual}. 
And we prefer additive attention to the multiplicative one, as it has experimentally shown to achieve a higher accuracy~\cite{schlemper2018}:

% ========================================
\begin{equation}
\begin{tabular}{rl}
$q^l_{att,i}$= & ${\bf W}_{a}^{T} \left( \sigma_1\,({\bf W}_x^T {\bf x}_i^l + {\bf W}_g^T {\bf g} + {\bf b}_{x} + {\bf b}_{g}\,)\,\right) + b_{a}$ \\
$\alpha^l$   = & $\sigma_2 (\, q^l_{att}({\bf x}_i^l\,,\, {\bf g}_i \,;\, \Theta_{att}) \,).$
\end{tabular}
\label{eq:alpha}
\end{equation}
%=========================================
\noindent
where $\sigma_1(x)$ is an element-wise nonlinear function, in particular, we use $= \sigma_1(x)= \max (x,0)$ and $\sigma_2(x)$ is a sigmoid activation function. Each AG is characterized by a set of parameters $\Theta_{att}$ containing the linear transformations ${\bf W}_x \in \mathbb{R}^{F_l \times F_{int}}$, ${\bf W}_g \in \mathbb{R}^{F_g \times F_{int}}$, ${\bf W}_{a} \in \mathbb{R}^{F_l \times 1}$, and bias terms $b_{a} \in \mathbb{R}$, ${\bf b}_{x}$, ${\bf b}_{g} \in \mathbb{R}^{F_{int}}$. AG parameters can be trained with the standard back-propagation updates together with other DispNet parameters. 

%===============================
\begin{figure*}[ht]
%    \begin{subfigure}
    \centering
    \includegraphics[width=0.32\textwidth]{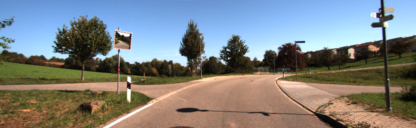} \,
    \includegraphics[width=0.32\linewidth]{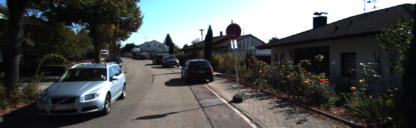} \,
    \includegraphics[width=0.32\linewidth]{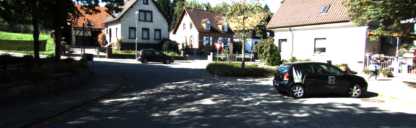}
%    \end{subfigure}
    
    \vspace{1mm}
    
%    \begin{subfigure}
    \centering
    \includegraphics[width=0.32\textwidth]{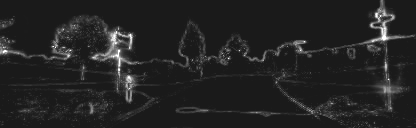} \,
    \includegraphics[width=0.32\linewidth]{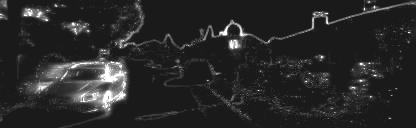} \,
    \includegraphics[width=0.32\linewidth]{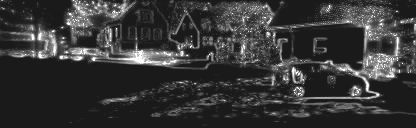}
%    \end{subfigure}
    \caption{Original images from KITTI dataset (top) and the corresponding attention coefficient maps (bottom).}
    \label{fig:att_coef}
\end{figure*}

%===============================
With attention gates integrated into DispNet, we modify the photometric loss in (\ref{eq:full}) accordingly, with attention coefficient $\alpha$ for pixel $p$ used instead of explainability value $E(p)$. Figure~\ref{fig:att_coef} in Section~\ref{sec:evaluation} visualizes the attention coefficients for three example images. It shows that the system pays less attention to moving objects, as well as to 2D edges and boundaries of regions with discontinuous depth values sensitive to the erroneous depth estimation.

\section{EVALUATION RESULTS}
\label{sec:evaluation}

\begin{table*}[h]
  \centering
%  \resizebox{0.9\textwidth}{!}{
  \begin{tabular}{l|c|c|c|c|c|c|c|c}
%  \toprule
  Method & Supervision  & \multicolumn{4}{|c}{Error metric} & \multicolumn{3}{|c}{Accuracy metric}\\ \hline
   & & Abs Rel & Sq Rel & RMSE  & RMSE log & $\delta < 1.25 $ & $\delta < 1.25^{2}$ & $\delta < 1.25^{3}$\tabularnewline \hline
%  \midrule
Eigen et al.~\cite{eigen2014depth} Coarse 
                        & Depth  & 0.214 & 1.605 & 6.563 & 0.292 & 0.673 & 0.884 & 0.957 \tabularnewline  \hline
Eigen et al.~\cite{eigen2014depth} Fine 
                        & Depth & 0.203 & 1.548 & 6.307 & 0.282 & 0.702 & 0.890 & 0.958 \tabularnewline  \hline
Liu et al.~\cite{Liu2016LearningFields} 
                        & Pose & 0.202 & 1.614 & 6.523 & 0.275 & 0.678 & 0.895 & 0.965 \tabularnewline \hline
Godard et al.~\cite{godard2016unsupervised} 
                        &No   & {\it 0.148} & 1.344 & 5.927 & 0.247  & 0.803 & 0.922 & 0.964 \tabularnewline \hline
Zhou et al.~\cite{zhou2017unsupervised} (w/o explainability) 
                        &No   & 0.221 & 2.226 & 7.527 & 0.294 &  0.676 & 0.885 & 0.954 \tabularnewline \hline
Zhou et al.~\cite{zhou2017unsupervised} (explainability) 
                        &No   & 0.208 & 1.768 & 6.856 & 0.283 &  0.678 &  0.885 & 0.957 \tabularnewline \hline
%\midrule
Almalioglu et al.~\cite{almalioglu2018ganvo} 
                        &No& \it{0.138} & {\it 1.155} & \bf{4.412} & {\it 0.232} & {\it 0.820} & {\it 0.939} & \bf{0.976} \tabularnewline \hline
Shen et al.~\cite{shen19icra} &No& 0.156 & 1.309 & 5.73 & 0.236 & 0.797 & 0.929 & 0.969 \tabularnewline \hline
Bian et al.~\cite{Bian19nips} &No& \bf{0.137} & \bf{1.089} & 6.439 & \bf{0.217} & \bf{0.830} &\bf{0.942} & {\it 0.975} \tabularnewline \hline
  \hline
\textrm{Ours} (BF)        &No& 0.213 & 1.849 & 6.781 & 0.288 & 0.679 & 0.887 & 0.957 \tabularnewline \hline
\textrm{Ours} (Attention) &No& 0.171 & 1.281 & 5.981 & 0.252 & 0.755 & 0.921 & 0.968 \tabularnewline \hline
\textrm{Ours} (BF+Attention)&No& 0.162 & \it{1.126} & {\it 5.284} & {\it 0.221} & \it{0.823} & \it{0.935} & \it{0.971} \tabularnewline \hline
%\bottomrule
\end{tabular}
%}
  \vspace{10pt}
  \caption{Single-view depth results on the KITTI dataset \cite{GeigerAreSuite} using the split of Eigen et al.~\cite{eigen2014depth} . Best and 2 runner-up results are shown in bold and italic, respectively.}
    \label{tab:kitti_eigen}
    \vspace{-10pt}
\end{table*}

In this section, we present the evaluation results of depth and ego-motion estimation, analyze them. We support our analysis with some visualizations, such as attention coefficients for masking pixels with likely corrupted image reconstruction.

%===========================================================
\subsection{Depth estimation}
\label{ssec:eva_depth}

We evaluated the depth estimation on publicly available KITTI Raw dataset. It contains 42,382 rectified stereo pairs from 61 scenes. The Image size is 1242$\times$375. For the comparison with the previous works, we adopt the test split proposed by Eigen et al.~\cite{eigen2014depth}. The split consists of 697 images that cover a total of 29 scenes. The remaining 32 scenes (23,488 images) are used for training/validation split with 22,600/888 images respectively. 
%=====================================
%\subsection{Quantitative results}
%\hide{%----------- eval metric are the same -------------- 
%\subsubsection{Evaluation metrics}
We adopt the evaluation metrics already used in previous works~\cite{eigen2014depth, GodardUnsupervisedConsistency, Garg2016UnsupervisedRescue}. They include the mean relative error (Abs Rel), the squared relative error (Sq Rel), the root mean squared error (RMSE), the mean $\log10$ error (RMSE log), and the accuracy with threshold $t$ where $t \in [1.25, 1.25^2, 1.25^3]$ (see~\cite{eigen2014depth} for more detail).

\hide{%-------------eval metrics in comments---------
Given $P$ the total number of pixels in the test set and $\hat{d}_i$, $d_i$ the estimated depth and ground truth depth values for pixel $i$, we have :
\begin{itemize}
    \item The mean relative error (Abs Rel): 
    $\frac{1}{P} \sum_{i=1}^{P} \frac{\parallel \hat{d}_i - d_i \parallel}{d_i}$, 

    \item The squared relative error (Sq Rel): 
    $\frac{1}{P} \sum_{i=1}^{P} \frac{\parallel \hat{d}_i - d_i \parallel^2}{d_i}$, 
    
    \item The root mean squared error (RMSE): $\sqrt{\frac{1}{P}\sum_{i=1}^P(\hat{d}_i - d_i)^2}$, 

    \item The mean $\log10$ error (RMSE log):
    $\sqrt{\frac{1}{P} \sum_{i=1}^{P} \parallel \log \hat{d}_i - \log d_i \parallel^2}$

    \item The accuracy with threshold $t$, \ie the percentage of
    $\hat{d}_i$ such that $\delta = \max (\frac{d_i}{\hat{d}_i},\frac{\hat{d}_i}{d_i}) < t$, where $t \in [1.25, 1.25^2, 1.25^3]$.
\end{itemize}
}%--------- put evaluation metrics in the comment  -------

We test our architecture presented in Section~\ref{sec:architecture}, in the baseline configuration, extended with Backward-Forward loss, attention gates, and both. Table~\ref{tab:kitti_eigen} reports our depth evaluation results and compares them to the state of art methods. 

As the table shows that our methods show state comparable, without using additional attention comparing to the baseline and it outperforms the supervised and most unsupervised techniques and shows performance comparable to the most recent methods~\cite{almalioglu2018ganvo,Bian19nips} which extend the baseline modules with additional modules and components.

%=========================================================
%\subsubsection
{\it Attention coefficients:} Figure~\ref{fig:att_coef} visualizes the effect of attention coefficients as masks for down-weighting image regions that get corrupted It actually visualizes the inverse attention, where white color refers to the low attention coefficient $\alpha_i$, thus having a lower weight; the black color refer to high values.

We can see in the figure, that low attention coefficients point to pixels that have a high probability to be corrupted. First of all, it concerns image regions corresponding to the moving objects. In addition, the region with discontinuous depth values is considered as corrupted as well. This often includes region boundaries. Thin objects like street light and sign poles are also down-weighted because of the high probability of depth discontinuity and corrupted image reconstruction.

These results support the hypothesis for using attention coefficients as a mask. It represents an alternative to the explainability module in~\cite{zhou2017unsupervised}. The region of interest is more likely the rigid object that the network will have more confidence to estimate their depth, it will be almost equal all over the object, like the segmentation problem. Also, the rigid objects are the most appropriate to estimate the change in position between frames, that is why it shut down the coefficient for the moving objects.

%%%%%%%%%%%%%%%%%%%%%%%%%%%%%%%%%%%%%%%%%%%%%%%%%%%%%%%%%%%%%%%%%%%%%%%%%%
\subsection{Pose estimation}
\label{ssec:pose}
%\subsection{KITTI VO Dataset}
We use the publicly available KITTI visual odometry dataset. The official split contains 11 driving sequences with ground truth poses obtained by GPS readings. We use sequence 09 and 10 to evaluate our approaches to align with previous SLAM-based works.
%that are based on SLAM techniques.

%\subsection{Quantitative results}
%\subsection{Evaluation metrics}
%\label{ssec:metrics}
We follow the previous works in using the {\it absolute trajectory error} (ATE) as evaluation metric. It measures the difference between points of the ground truth and the predicted trajectory. Using timestamps to associate the ground truth poses with the corresponding predicted poses, we compute the difference between each pair of poses, and output the mean and standard deviation.
%\begin{itemize}
%    \item Average Trajectory Error (ATE): 
%    $\frac{1}{T} \sum_{t=1}^{T} \parallel \hat{P}_t - P_t \parallel$, where $T$ is total number %of timestamps, $\hat{P}_t$ and $P_t$ are the estimated and ground truth pose respectively.
%\end{itemize}
%To align with the previous works and for 
%For the sake of comparison with the previous works, 
%instead of using it the same way as in the equation above, 
%we calculate the average over every $k=$ consecutive frames fed to the poss module.

%=======================================================
% \subsubsection{Evaluation results}
% \label{ssec:ate-eval}

Table~\ref{tab:ate} reports the evaluation results of the Backward-Forward and attention modules, separately and jointly, and compare them to the previous works. Both extensions improve the baseline, the attention module performs well. When coupled with the BF, the attention boosts the performance of PoseNet training to have a consistent ego-motion estimation and outperforms all the state of the art methods~\cite{zhou2017unsupervised,mahjourian2018unsupervised,almalioglu2018ganvo,shen19icra}, which use additional models or pixel masks, thus increasing considerably the model size.

% Figure~\ref{fig:09} visualizes the pose estimation for the test sequence 09. For each degree of freedom, the translation ($x,y,z$) and rotation (roll, pitch, yaw), it compares the estimation by different methods to the ground truth.

%-----------------------------------------------------

%-----------------------------------------------------
\begin{figure*}[h]
\centering
%   \begin{subfigure}
    \includegraphics[width=0.48\linewidth]{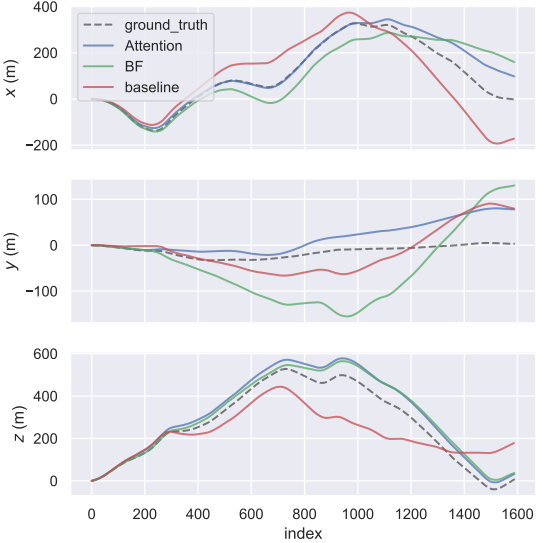}
    % \caption{a)Translation coordinates}
%   \end{subfigure}
%   \begin{subfigure}
    \includegraphics[width=0.48\linewidth]{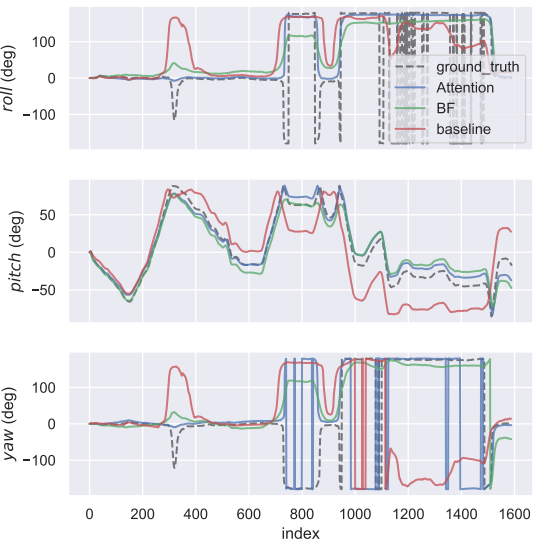}
    % \caption{b)Rotation coordinates}
%   \end{subfigure}
  \caption{Sequence 09: Comparing the pose estimation to the ground truth, for each degree of freedom, including the translation $x,y,z$ (left) and rotation roll, pitch, yaw (right).} 
  \label{fig:09}
\end{figure*}

%----------------------------------------------------

%-----------------------------------------------------
Figure~\ref{fig:09} shows the 6-DoF estimated throughout the trip of the sequence 09. Our improvements are more robust compared to the baseline except in the altitude (y). The figure unveils another important issue. It demonstrates the oscillation of roll and yaw values and their discontinuity when put in $[-\pi,\pi]$ interval while orientation changes are continuous in the real world. A recent analysis of 3D and $n$-dimensional rotations~\cite{zhou19continuity} shows that discontinuous orientation representations make them difficult for neural networks to learn. Therefore, it might be a subject of deeper analysis and a need of replacing quaternions with an alternative, continuous representation.

%---------------------------------------
\begin{table}[h]
\centering
%\scalebox{1.00}
%{
\begin{tabular}{l|c|c}
%\toprule
\textrm{Method} & \textrm{Seq. $09$} & \textrm{Seq. $10$} \tabularnewline \hline
%\midrule
\textrm{ORB-SLAM (full)} & $0.014 \pm 0.008$ & $0.012 \pm 0.011$  \tabularnewline \hline
\textrm{ORB-SLAM (short)} & $0.064 \pm 0.141$ & $0.064 \pm 0.130$ \tabularnewline \hline
\textrm{Zhou et al.~\cite{zhou2017unsupervised} (baseline)}  & $0.016 \pm 0.009$ & $0.013 \pm 0.009$ \tabularnewline \hline
\textrm{Mahjourian et al.~\cite{mahjourian2018unsupervised}} & $0.013 \pm 0.010$ & $0.012 \pm 0.011$ \tabularnewline \hline 
\textrm{Almalioglu et al.~\cite{almalioglu2018ganvo}}& $0.009 \pm 0.005$ & $0.010 \pm 0.0013$ \tabularnewline \hline
\textrm{Shen et al.~\cite{shen19icra}}& $0.0089 \pm 0.0054$ & $0.0084 \pm 0.0071$ \tabularnewline \hline \hline 
%\textbf{Ours (GAN-based)} & $0.019  \pm 0.016$ & $0.015 \pm 0.014$ \tabularnewline
\textrm{Ours (BF)}          &$0.0101 \pm 0.0065$ & $0.0091 \pm 0.0069$ \tabularnewline \hline
\textrm{Ours (Attention)}   &$0.0108 \pm 0.0062$ & $0.0082 \pm 0.0063$ \tabularnewline \hline
\textrm{Ours (BF+Attention)}&\bf{0.0087 $\pm$ 0.0054}&\bf{0.0078 $\pm$ 0.0061}
\tabularnewline \hline
% \midrule
% \textbf{Ours (GAN-based), k=3} & $0.012 \pm 0.006$ & $0.010 \pm 0.007$ \tabularnewline
% \textbf{Ours (Attention), k=3} & $0.009 \pm 0.005$ & $0.009 \pm 0.007$ \tabularnewline
%\bottomrule
\end{tabular}
%}
\vspace{0.1cm}
\caption{Absolute Trajectory Error (ATE) on the KITTI odometry split averaged over all frame snippets (lower is better).}
\label{tab:ate}
\end{table}

\subsection{Model size and the training time}
Unlike the explainability, the attention gates do not require an additional networks, they are layers integrated into the existing depth network. Our architecture for DEM evaluation requires fewer parameters in the model and shows a faster training time than the state of the art methods. Indeed, adding the BF loss has a negligible impact on the training time with respect to the baseline. Adding attention gates increases the model size by 5-10\% and training time by 10-25\%. For the comparison, the training time of additional semantic segmentation~\cite{mahjourian2018unsupervised} or GAN module~\cite{almalioglu2018ganvo} doubles the model size and requires 4-10 more time to train the models. 
%%%%%%%%%%%%%%%%%%%%%%%%%%%%%%%%%%%%%%%%%%%%%%%%%%%%%%%%%%%%%%%%%%%%%%%%%%%%%%%%

\section{CONCLUSIONS}
\label{sec:conclusion}

We have presented two extensions of the baseline architecture for the depth and pose estimation tasks, all aimed to improve the performance in the self-supervised setting. Adding backward-forward consistency loss to the training process allowed to boost the performance. Our method follows one of the current trends forcing the learned models to respect the geometric principles but adding penalties for any consistency violation. This idea opens a possibility to explore and impose more geometric constraints on the learned models, this might further improve the accuracy. We have shown the effectiveness of attention gates integrated in the depth module of DEM estimation. It demonstrates that the attention principle can be expanded to the navigation tasks. The attention gates help identify corrupted image regions where the static world assumption is violated. Besides, the attention model can be explored as masking coefficients in multiple different ways, it represents a strong alternative to the explainability network in the baseline architecture. 

%%%%%%%%%%%%%%%%%%%%%%%%%%%%%%%%%%%%%%%%%%%%%%%%%%%%%%%%%%%%%%%%%%%%%%%%%%%%%%%%

\bibliographystyle{plain}
\bibliography{bibfile,references}
\end{document}